\documentclass[]{spie}  

\usepackage{amsmath,amsfonts,amssymb}
\usepackage{graphicx}
\usepackage[colorlinks=true, allcolors=blue]{hyperref}

\usepackage{booktabs}
\usepackage{comment}
\usepackage{dirtytalk}
\usepackage{amsfonts,amssymb}
\usepackage{adjustbox}
\usepackage{subcaption}
\usepackage{mathtools}
\usepackage{hyperref}
\usepackage{array, caption, floatrow, tabularx, makecell, booktabs}%
\setcellgapes{1pt}

\title{Towards real-time 6D pose estimation of objects in single-view cone-beam X-ray}

\author[ab]{Christiaan G.A. Viviers}
\author[a]{Jo\"{e}l de Bruijn}
\author[b]{Lena Filatova}
\author[a]{Peter H.N. de With}
\author[a]{Fons van der Sommen}

\affil[a]{Eindhoven University of Technology, 5612 AP, Eindhoven, the Netherlands;}
\affil[b]{Philips IGT, 5684 PC, Best, the Netherlands;}

\authorinfo{Further author information: Christiaan G.A. Viviers: E-mail: c.g.a.viviers@tue.nl: Telephone: +31 (0)6 206 60171}

\begin{document} 
\maketitle

\begin{abstract}
Deep learning-based pose estimation algorithms can successfully estimate the pose of objects in an image, especially in the field of color images. 6D Object pose estimation based on deep learning models for X-ray images often use custom architectures that employ extensive CAD models and simulated data for training purposes. Recent RGB-based methods opt to solve pose estimation problems using small datasets, making them more attractive for the X-ray domain where medical data is scarcely available. We refine an existing RGB-based model (SingleShotPose) to estimate the 6D pose of a marked cube from grayscale X-ray images by creating a generic solution trained on only real X-ray data and adjusted for X-ray acquisition geometry. The model regresses 2D control points and calculates the pose through 2D/3D correspondences using Perspective-n-Point (PnP), allowing a single trained model to be used across all supporting cone-beam-based X-ray geometries. Since modern X-ray systems continuously adjust acquisition parameters during a procedure, it is essential for such a pose estimation network to consider these parameters in order to be deployed successfully and find a real use case. With a 5-cm/5-degree accuracy of 93\% and an average 3D rotation error of 2.2 degrees, the results of the proposed approach are comparable with state-of-the-art alternatives, while requiring significantly less real training examples and being applicable in real-time applications.
\end{abstract}

\keywords{6D pose estimation,  object detection, deep learning, X-ray projection model}

\section{INTRODUCTION}
\label{sec:intro}  
Precise estimation of the pose (position and orientation) of an object from an image is of high importance in various applications, including the medical domain~\cite{Hatt2016, i3PosNet}. Impressive advancements have recently been  made in the application of deep learning Convolutional Neural Networks (CNNs), to solve the 6D pose estimation problem. The majority of these advances has been in the domain of color and depth images, driven by the automotive (autonomous driving) and robotics industry, while the domain of medical image analysis has seen fewer developments. This can be largely attributed to the limited availability --due to privacy restrictions-- of large medical image datasets suitable for training such deep learning-based pose estimation models. Custom architectures have been proposed for object pose prediction in medical images, largely relying on simulated datasets for training purposes, then followed by fine-tuning the model using the real data~\cite{xrayPosNet, i3PosNet}. Two popular strategies for training neural networks in a supervised manner, are (1)~through randomly initializing the network weights, thus \say{training from scratch}, and (2)~by pre-training the model on a related task followed by further refinement on the target task. Training a supervised model from scratch typically requires a very large labeled dataset to obtain state-of-the-art (SOTA) and accurate results. Unfortunately, in the medical domain such datasets are scarce, due to privacy restrictions and the expertise required for the labeling process. As a result, transfer learning has become a popular approach when developing deep learning models for medical image analysis. In addition to this technique, training deep pose estimation networks require labeled datasets that contain object class labels, projection coordinates and, depending on the model of choice, a segmentation mask as well as a 3D model of the object. Creating such a dataset can be extremely time-consuming. It is possible to simulate the objects of interest to develop the required labeled dataset, but this approach often leaves a domain gap and yields lower performance on the final real test dataset~\cite{xrayPosNet}. As mentioned, the simulation process itself has its drawbacks and thus, a transfer learning approach is proposed.

To utilize advancements in deep pose estimation networks developed specifically for the RGB-domain and then exploit real data, we employ SingleShotPose~\cite{yolo6D}, an RGB-based 6D object-pose prediction model. The SingleShotPose model was designed for simultaneous single-shot object detection and 6D pose prediction from an RGB image. This is realized by directly predicting the 2D image locations of the projected vertices of the object’s 3D bounding box. Using a Perspective-n-Point (PnP) algorithm and known acquisition parameters, the 6D pose of an object can be then estimated. Since the SingleShotPose model only predicts 2D image locations and does not rely on learned acquisition parameters, the PnP solver can employ varying acquisition geometries to solve the 6D pose in a generic way. This allows for a single trained model to be used across all supporting cone-beam-based X-ray geometries.

X-ray imaging modalities often contain only a single grayscale channel. It is possible to mimic the RGB structure of natural images through copying of the grayscale amplitudes into the individual channels of a pseudo-color image. However, the stacked grayscale image does not contain RGB color information and therefore, it is expected that the initial network layers for filtering learned from color images remain underutilized. To address this, we have pre-trained the part of the SingleShotPose model for feature extraction on a grayscale (Luma-transformed~\cite{luma}) ImageNet classification task. We have observed that the classification model's accuracy only experiences a minor degradation ($2\%$), notwithstanding the lack of color within these images. The pre-trained weights serve as a starting point for the SingleShotPose model when being trained for the domain-specific task of predicting the pose of a marked cube in X-ray images.

\section{RELATED WORK}
\label{sec:relatedwork}
\subsection{Pose estimation using deep neural networks}
\label{ssec:PoseEstimation}
There are several points to consider when choosing a model for the task of object pose estimation in X-ray images. Firstly, due to the nature of the acquisition process of X-ray images, RGB-based models that exploit textures of objects (for example,  DPOD~\cite{DPOD}) can be disregarded, since textures are less prevalent and subject to a considerable change in appearance due to their transparency observed in X-ray. Secondly, methods such as SSD6D~\cite{SSD6D} exploit depth information from RGB-D data, which is not available in typical medical imaging apparatus. Thus, these types of models can be omitted. A third aspect to consider is inference speed. RGB-based pose estimation models typically have two layouts: single-stage and multi-stage methods. Multi-stage methods, such as BB8~\cite{BB8}, tend to be rather slow, often processing only a few frames per second. This reduces their relevance in any real-time analysis applications. There are two popular implementations of single-stage methods. Methods such as EfficientPose~\cite{EfficientPose} regress the object pose directly from the image. The other popular single-stage implementation entails a deep network that predicts control points in the 2D image and computes the object pose through 2D/3D correspondences. Directly regressing the pose from the image assumes the acquisition camera intrinsic parameters are static, unless these parameters are added as input to the prediction network. Solving the pose through 2D/3D correspondences allows for different acquisition models to be used, of which the pinhole camera model~\cite{pinhole} is quite popular and the pose can then easily be solved via PnP~\cite{PnP} methods. 

\subsection{Pose estimation in X-ray}
\label{ssec:PEinXRAY}
To the best of our knowledge, pose estimation of objects in X-ray images via deep learning-based methods have only been attempted in three cases. K\"{u}gler \textit{et al.}~\cite{i3PosNet} reported a VGG-based pose estimation network to infer the position and orientation of instruments from images. The network uses localized patches and outputs pseudo-landmarks (control points). The pose is then reconstructed from the pseudo-landmarks by geometric considerations. Their implementation was fully trained on simulated data and finally tested on real X-ray images. Presenti \textit{et al.}~\cite{Presenti2020} used a ResNet-50-based architecture to directly estimate the rotation (3 Euler angles) of an object from simulated X-ray images for application in CT. Lastly, X-ray PoseNet~\cite{xrayPosNet} is a custom CNN that also directly estimates the pose from the X-ray image. It regresses the translation (3 degrees) and rotation (4 quaternions) and is trained on both simulated and real X-ray images. It is also noteworthy that these methods assume fixed calibration/acquisition parameters and they are all trained using simulated images.

In our case, we concentrate on developing a generic model that is applicable to multiple geometries that can change at run-time, while avoiding the extensive simulation process that often does not translate to the real environment effectively.

\section{Methodology}
\label{sec:methodology}

\subsection{X-ray marked cube dataset}
\label{ssec:CubeDataSet}
\begin{figure}\RawFloats\CenterFloatBoxes
\begin{floatrow}
\ffigbox[\FBwidth]
{\includegraphics[width=3.5cm, trim={1cm 1cm 1cm 1cm},clip]{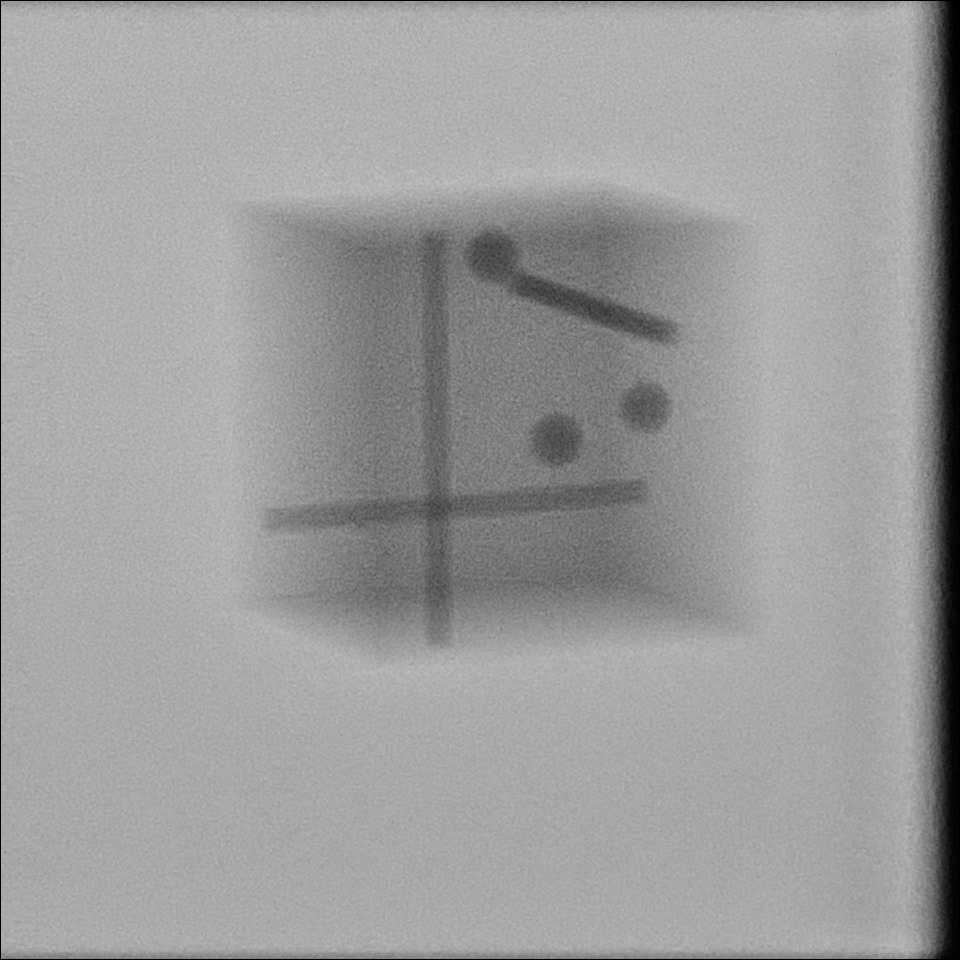}
\includegraphics[width=3.5 cm, trim={0.1cm 0.1cm 0.1cm 0.1cm},clip]{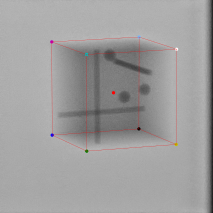}}
{\caption{Zoomed example of an X-ray image in the marked cube dataset and its corresponding label.}\label{fig:Cube_dataset_example}}
\ttabbox%
{\scriptsize{\begin{tabularx}{0.45\textwidth}{l *{2}{>{\centering\arraybackslash}X}}
        \toprule
        \rule[-1ex]{0pt}{3.5ex}  Rotation  & Translation (mm)\\
        \midrule
        \rule[-1ex]{0pt}{3.5ex}  $r_{z} \in [0^{\circ}, 360^{\circ}]$, $r_{y} \in [0^{\circ}, 360^{\circ}]$ &  $ t_{z} = 700 \pm 40 $, $ t_{y} = 0 \pm 40 $  \\ 	
        \rule[-1ex]{0pt}{3.5ex}  $r_{x} \in [-45^{\circ},45^{\circ}], r_{x} \in [135^{\circ}, 225^{\circ}]$ &  $t_{x} = 0 \pm 40 $   \\
        \bottomrule
        \rule[-1ex]{0pt}{3.5ex}  SID  (mm)  & FOV (mm) diagonal\\
        \midrule
        \rule[-1ex]{0pt}{3.5ex}  $ [1100.0, 1230.0]$   & $[156 , 297]$ \\	
        \bottomrule 
      \end{tabularx}}}
{\caption{Parameters for constructing the marked cube dataset.}\label{tab:cubedataset}}
\end{floatrow}
\end{figure}

We constructed the marked cube X-ray dataset to evaluate the performance 6D pose estimation model on X-ray images, captured with a C-arm-based X-ray system (Azurion, Philips IGT, Best, Netherlands) and referred to as real X-ray data (example image and label in Figure~\ref{fig:Cube_dataset_example}). The 30$\times$30$\times$30-mm perspex cube is embedded with metal markers. Table~\ref{tab:cubedataset} contains the acquisition parameters and the range in which they vary in the constructed dataset. The dataset consists of 2,042 manually annotated 8-bit 960$\times$960-pixel grayscale images, divided over a 80/20 train/validation split. 
\subsection{Cone beam X-ray geometry model}
\label{ssec:ConeBeam}
\begin{figure}[ht!]
\begin{minipage}{0.5\textwidth}
\centering
  \includegraphics[width=8cm, trim={0.0cm 0.5cm 0.0cm 0.0cm},clip]{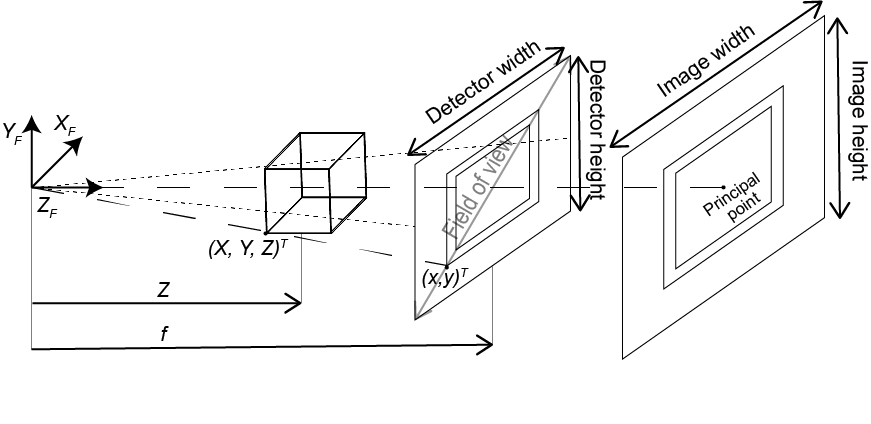}
  \caption{X-ray projection model.}
  \label{fig:projection model}
\end{minipage}
\begin{minipage}{0.5\textwidth}
\footnotesize
\begin{equation} \label{eq:projection_model}
\lambda \begin{bmatrix*}[c] u \\ v \\ 1 \end{bmatrix*} = \begin{bmatrix*} \textbf{K}  & \mathbf{O}_3 \end{bmatrix*}
    \begin{bmatrix*}[c] 
    \mathbf{R} & \mathbf{C} \\ \mathbf{O}_3^T & 1 
    \end{bmatrix*}
    \begin{bmatrix*}[c]
    X_W \\ Y_W \\ Z_W \\ 1 
    \end{bmatrix*}
\end{equation}
\begin{equation}
\mathbf{K}=
\begin{bmatrix*}[c]
k_{u}f     & 0    & k_{u}x_{0} \\
0   & -k_{v}f  & k_{v}y_{0} \\
0   & 0    & 1 
\end{bmatrix*}
\label{eqn:K_params}
\end{equation}
\null
\par\xdef\tpd{\the\prevdepth}
\end{minipage}
\end{figure}

Modern X-ray systems allow for the X-ray acquisition parameters to be changed at run-time. Acquisition parameters such as the field of view (FOV) and source-image distance (SID), i.e. the focal length, can be changed by the user to improve the view on the area of interest. The X-ray acquisition model (Figure~\ref{fig:projection model}) explained by Equation~\ref{eq:projection_model} captures these parameters. 

In Equation~\ref{eq:projection_model}, $\lambda$ is a scaling factor and $u$ and $v$ denote the projected 2D coordinates of the 3D point. Matrix $\mathbf{K}$ represents the system intrinsic parameters and is further expanded in Equation~\ref{eqn:K_params}. Matrix $\mathbf{R}$ is the $3 \times 3$ rotation matrix and $\mathbf{C}$ is a $3 \times 1$ vector, representing the rotation and translation of the object with respect to the imaging source and $\mathbf{O}_{n}$ is a $n \times 1$ vector of zeroes. The last vector consists of the 3D object coordinates in the world frame. 
Matrix $\mathbf{K}$ is often referred to as the intrinsic parameters and calibration matrix. Let $k_{u}$ and $k_{v}$ be the horizontal and vertical density of pixels. In the matrix, $f$ refers to the focal length or SID in X-ray systems. Parameters $x_{0}$ and $y_{0} $ account for the offset of the principal point to the detector center. As mentioned before, modern X-ray systems allow for the focal length and the imaging field of view to be changed at run-time by the user (resulting in variable effective detector sizes and thus a different offset to the image center). Changing these parameters results in a different projection of the object of interest onto the detector, even though the object's pose has not changed. 
\subsection{Pose estimation model}
\label{ssec:Model}
The SingleShotPose model~\cite{yolo6D} performs object detection and regresses the 2D image coordinates from an image, more specifically the object’s centroid and the corners of its 3D bounding box. The model does not require depth information and it is a single-staged model, thereby enabling short inference times. The model uses 2D object-coordinate predictions in combination with the PnP algorithm, which uses a set of camera-intrinsic parameters, and the 3D vertices' coordinates to calculate the rotation and translation of an object. In X-ray imaging systems, these intrinsic parameters (shown in Equation~\ref{eqn:K_params}) are known and changes at run-time. Given these aforementioned properties, the SingleShotPose model is expected to be applicable, with minor adjustments, to medical X-ray data analysis and 6D object pose estimation from a single image.
\subsection{Training and preparing the model}
\label{ssec:training}
For the purpose of predicting the 6D pose of the marked cube in our X-ray dataset, we adjust the SingleShotPose model. We change the model's input kernel by reducing its depth from 3 to 1, aiming to adapt it for single-channel grayscale images. The model is then trained, using a similar training strategy as described in the original paper~\cite{yolo6D}, except that the network parameters are initialized by training the base network (Darknet-19 448$\times$488 used in Yolo V2~\cite{YOLOv2}) on the now grayscale ImageNet classification task. Similar to the work of Xie \textit{et al.}~\cite{Xie_2018_ECCV_Workshops}, it can be observed that the classification accuracy only drops slightly, even in the absence of color (Table~\ref{tab:pretraining}). We then train the complete model on the marked cube dataset for the detection and 6D pose prediction task. We employ extensive data augmentation by randomly scaling (up to 40\%) and translating the image by a factor of up to 20\% of the image size. Using an object mask, we also blend the background (by a random amount) with a grayscale-transformed random image from the PASCAL VOC dataset. In our case, we train using an ADAM optimizer for 500 epochs starting at a learning rate of 0.001 and divide the learning rate by 10 every 80 epochs.  The SingleShotPose is adjusted for this research and implemented in PyTorch 1.7.0. All experiments are carried out on an NVIDIA RTX~2080Ti GPU.
\begin{table}[ht]
\centering
\begin{tabular}{llcc} 
\toprule
\rule[-1ex]{0pt}{3.5ex}  Dataset & Model  & Top-1 (\%)  & Top-5 (\%) \\
\midrule
\rule[-1ex]{0pt}{3.5ex}  ImageNet & Darknet19 & 72.9 & 91.2  \\ 	
\rule[-1ex]{0pt}{3.5ex}  ImageNet & Darknet19\_488 & 76.4 & 93.5  \\ 	
\rule[-1ex]{0pt}{3.5ex}  Grey ImageNet & Darknet19 & 70.1 &  89.8   \\
\rule[-1ex]{0pt}{3.5ex}  Grey ImageNet & Darknet19\_488 & 74.0  & 91.8   \\
\bottomrule 
\end{tabular}
\caption{Darknet training results on luma grayscale-transform ImageNet classification task compared to J. Redmon and A. Farhadi~\cite{YOLOv2}.}
\label{tab:pretraining}
\end{table}
\section{EXPERIMENTAL RESULTS}
\label{sec:experimentalresults}
\subsection{Evaluation metrics}
\label{ssec:metrics}

We use the 2D reprojection error,  the commonlyused 5-cm/5-degree metric and the ADD metric to evaluate the 6D pose accuracy. Additionally, we report the pose prediction speed (FPS) of the model. Following Brachmannet \textit{et al.}~\cite{Brachmann_2016_CVPR}, a 2D-projection prediction is considered correct when the average distance between ground-truth projection and the predicted 2D-projected vertices of the object model is smaller than 5 pixels. Given that our images are 3 times larger than the LineMod (640$\times$480) dataset (on which the metric was established), we also evaluate the 2D projection error at different thresholds. The 5-cm/5-deg metric considers a prediction correct if the error is below 5~cm and 5~degrees. Using the ADD metric, a predicted 6D pose is considered correct if the average 3D point distance is smaller than 10\% of the object diameter from the ground-truth point.
\begin{table}[ht]
\centering
\normalsize {
\begin{tabular}{cccc} 
\toprule
\rule[-1ex]{0pt}{3.5ex}  Input Res. & 2D Reproject (5px)  & 5cm/5deg & FPS\\
\midrule
\rule[-1ex]{0pt}{3.5ex}  $960 \times 960$ & 13.7\%  & \textbf{93.2\%} & 35.7 \\ 
\rule[-1ex]{0pt}{3.5ex}  $800 \times 800$ & \textbf{17.6\%}  & 90.2\% & 45.5\\ 
\rule[-1ex]{0pt}{3.5ex}  $672 \times 672$ & 17.4\%  & 78.2\% & \textbf{52.6} \\
\bottomrule 
\end{tabular}
}
\caption{Results obtained at different image input resolutions.}
\label{tab:5cm5degresults}
\end{table}

\subsection{Object pose estimation}
\label{ssec:analysis}

\begin{figure}[ht]\TopFloatBoxes
\begin{floatrow}[2]
\floatrowsep
        \ffigbox[\FBwidth]
        {\caption{Model predictions (blue) and ground truth (green).}\label{fig:example_pred}}
        {
        \includegraphics[width=4cm, trim={1cm 1cm 1cm 1cm},clip]{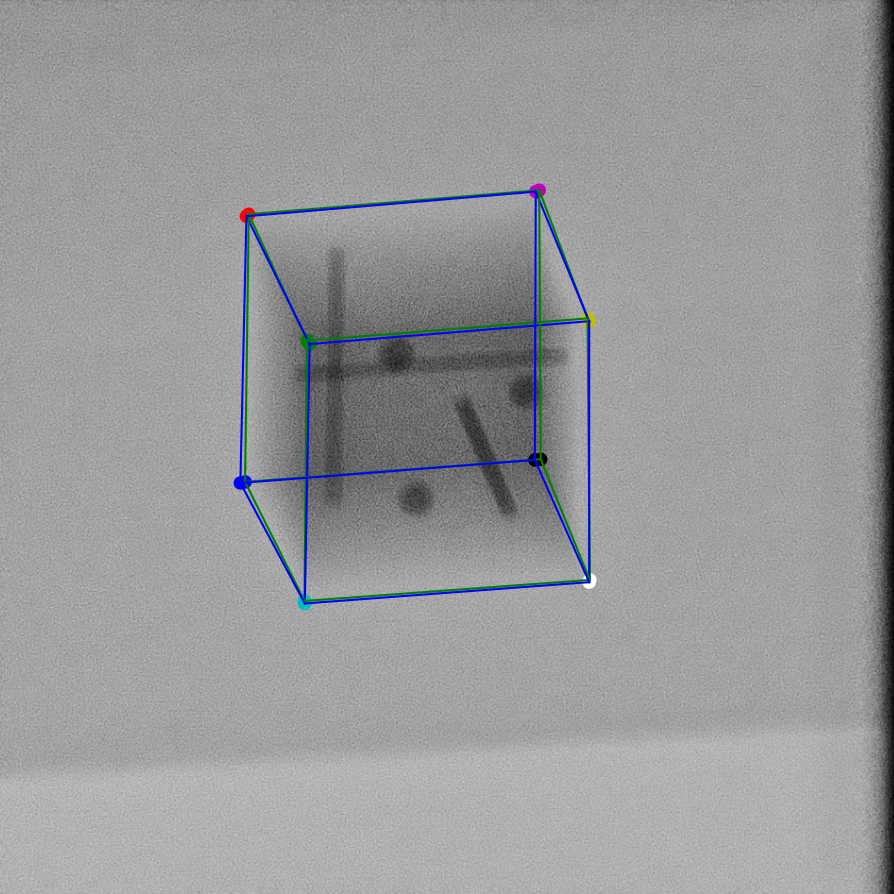}
        \includegraphics[width=4cm, trim={1cm 1cm 1cm 1cm},clip]{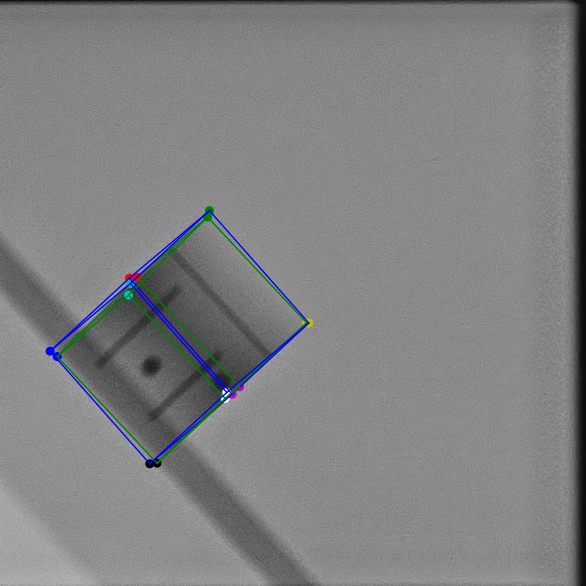}}
  \ttabbox[\Xhsize]
    {\small{
    \begin{tabularx}{0.5\textwidth}{ l *{4}{>{\centering\arraybackslash}X}}
      \toprule
        \rule[-1ex]{0pt}{3.5ex}  & 5~pixels  & 10~pixels  & 15~pixels\\
        \hline
        \rule[-1ex]{0pt}{3.5ex} 2D Acc   & 13.7\% & 67.7\% & 91.2\%  \\ 
        \bottomrule
        \rule[-1ex]{0pt}{3.5ex}   & ADD (10\%) & ADD (50\%) & ADD (100\%)\\
        \hline
        \rule[-1ex]{0pt}{3.5ex} 3D Acc & 10.0 \% & 51.6\% & 81.7\%\\ 
        \bottomrule 
        \rule[-1ex]{0pt}{3.5ex}   & 2D Pixel & 3D Angle (deg.) & 3D Transl. (mm)\\
        \hline
        \rule[-1ex]{0pt}{3.5ex} Error & $9.2 \pm 4.7$ & $2.2 \pm  1.2$ & $17.7 \pm 14.8$\\ 
        \bottomrule 
      \end{tabularx}}}
    {\caption{2D projection accuracy, ADD metric and average error evaluated on 960$\times$960 pixel images.}
     \label{tab:2Dprojection_results}}%
        
\end{floatrow}%
\end{figure}

The  results  of  our  model on the marked cube test dataset are shown in Table~\ref{tab:5cm5degresults} and Table~\ref{tab:2Dprojection_results}.  The green 3D bounding boxes in Figure~\ref{fig:example_pred} visualize ground-truth poses while our estimated pose is represented by blue boxes. The corners are represented by consistent colors. Considering the X-ray domain and high input resolution, we observe that the model has success in predicting the 2D bounding-box projection coordinates by the 2D reprojection metric. Since the pose is calculated from the 2D control-point predictions, the variance in accuracy of the 2D estimations directly translate to the 6D pose. We achieve a high 3D rotation accuracy. The control-point estimations are accurate in relative orientation, whereas some variance occurs in overall scale. Our results are obtained without the use of extensive CAD models and simulated data. Since the trained model accepts input images of different resolutions and the PnP solver accounts for the X-ray system acquisition geometry, this single trained model can be used across multiple systems. The SingleShotPose model executes at low inference times (27.6 ms), which makes it valuable for real-time applications.

\section{CONCLUSION}
\label{sec:conclusion}
This research shows that advancements in RGB-based pose-estimation CNN models are applicable to a broad range of cases, easily stretching into the medical X-ray imaging domain. In the absence of medical datasets, our proposed cross-domain transfer learning offers a more efficient alternative, which is demonstrated by the adaptation from the RGB-to-grayscale case and then finally going to the X-ray domain. The chosen deep learning network, SingleShotPose, regresses the bounding-box coordinates of the object of interest in the 2D image. Using the X-ray cone-beam model, the object pose is then calculated through 2D/3D correspondences and a PnP solver. This proposed approach achieves a \say{one model fitting all X-ray cone-beam geometries}-method by explicitly excluding the projection geometry-related parameters from the trained network. Although it is limited in its ability to accurately regress the 2D projected coordinates, we conjecture a more expressive deep network will improve this in future work. Finally, our generic approach enables quite high pose accuracy from a single X-ray image, especially in the 3D orientation, whilst being applicable in real-time applications.
\newpage
\bibliography{report} 
\bibliographystyle{spiebib} 

\end{document}